%% file: mg.tex
\documentclass[11pt,a4paper]{article}
\pdfoutput=1
\usepackage[hyperref]{acl2019}
\usepackage{times}
\usepackage{soul}
\usepackage{url}
\usepackage{algorithm}
\usepackage{algorithmic}
\usepackage{balance}
\newcommand{\ModelName}{\textsc{MGNER}}
\usepackage{amsmath,amsthm,mathtools}
\usepackage{latexsym,graphicx,amssymb,multirow,booktabs}

\usepackage{hyperref}
\usepackage{url}
\usepackage{xcolor}
\usepackage{soul}
\usepackage{enumitem}

\aclfinalcopy % Uncomment this line for the final submission
\def\aclpaperid{987} %  Enter the acl Paper ID here

\newcommand{\multi}{Multi-Grained}
\title{{\multi} Named Entity Recognition}

\author{Congying Xia{$^{1,5}$},~Chenwei Zhang{$^1$}, ~Tao Yang$^2$, ~Yaliang Li$^{3}$\thanks{~ Work was done when the author Yaliang Li was at Tencent America.},\\
\textbf{Nan Du}$^{2}$\textbf{,} \textbf{Xian Wu}$^{2}$\textbf{,} \textbf{Wei Fan}$^{2}$\textbf{,} \textbf{Fenglong Ma}$^{4}$\textbf{,} \textbf{Philip Yu}{$^{1,5}$}\\
  {$^1$University of Illinois at Chicago, Chicago, IL, USA} \\
  {$^2$Tencent Medical AI Lab, Palo Alto, CA, USA};
    {$^3$Alibaba Group, Bellevue, WA, USA}\\
    {$^4$University at Buffalo, Buffalo, NY, USA};
   {$^5$Zhejiang Lab, Hangzhou, China}\\
  {\tt \{cxia8,czhang99,psyu\}@uic.edu};  
   {\tt yaliang.li@alibaba-inc.com} \\
  {\tt \{tytaoyang,kevinxwu,davidwfan\}@tencent.com}\\
 {\tt nandu2048@gmail.com};
  {\tt fenglong@buffalo.edu} \\
}

\date{}

\begin{document}
\maketitle
\begin{abstract}
This paper presents a novel framework, {\ModelName}, for {\multi} Named Entity Recognition where multiple entities or entity mentions in a sentence could be non-overlapping or totally nested. 
Different from traditional approaches regarding NER as a sequential labeling task and annotate entities consecutively, {\ModelName} detects and recognizes entities on multiple granularities: it is able to recognize named entities without explicitly assuming non-overlapping or totally nested structures.
{\ModelName} consists of a Detector that examines all possible word segments and a Classifier that categorizes entities.
In addition, contextual information and a self-attention mechanism are utilized throughout the framework to improve the NER performance.
Experimental results show that {\ModelName} outperforms current state-of-the-art baselines up to 4.4\% in terms of the F1 score among nested/non-overlapping NER tasks.
\end{abstract}

\input{1intro.tex}
\input{2related.tex}
\input{3model.tex}
\input{4exp.tex}
\input{5con.tex}
\input{6ack.tex}

\bibliographystyle{acl_natbib}
\balance
\bibliography{acl2019}

\end{document}

%% file: 1intro.tex
\section{Introduction}
Effectively identifying meaningful entities or entity mentions from the raw text plays a crucial part in understanding the semantic meanings of natural language. Such a process is usually known as Named Entity Recognition (NER) and it is one of the fundamental tasks in natural language processing (NLP). A typical NER system takes an utterance as the input and outputs identified entities, such as person names, locations, and organizations. The extracted named entities can benefit various subsequent NLP tasks, including syntactic parsing \cite{koo2010efficient}, question answering \cite{krishnamurthy2015learning} and relation extraction \cite{lao2010relational}. However, accurately recognizing representative entities in natural language remains challenging.

\begin{figure}[bpth!]
    \centering
    \includegraphics[width=0.8\linewidth]{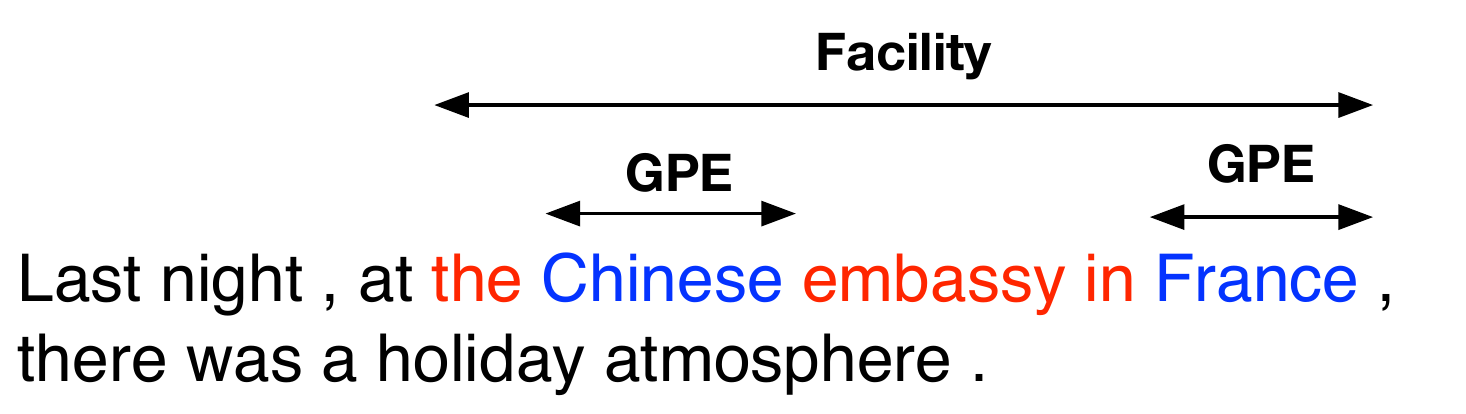}
    \caption{An example from the ACE-2004 dataset \cite{doddington2004automatic} in which two GPEs (Geographical Entities) are nested in a Facility Entity. } 
    \label{fig:multi}
\end{figure}

Previous works treat NER as a sequence labeling problem. For example, \citet{lample2016neural} achieve a decent performance on NER by incorporating deep recurrent neural networks (RNNs) with conditional random field (CRF) \cite{lafferty2001conditional}.
However, a critical problem that arises by treating NER as a sequence labeling task is that it only recognizes non-overlapping entities in a single, sequential scan on the raw text; it fails to detect nested named entities which are embedded in longer entity mentions, as illustrated in Figure \ref{fig:multi}. 

Due to the semantic structures within natural language, nested entities can be ubiquitous: {\it e.g.} 47\% of the entities in the test split of ACE-2004 \cite{doddington2004automatic} dataset overlap with other entities, and 42\% of the sentences contain nested entities.
Various approaches \cite{alex2007recognising, lu2015joint, katiyar2018nested, lu2017mention, wang2018neural} have been proposed in the past decade to extract nested named entities.
However, these models are designed explicitly for recognizing nested named entities. They usually do not perform well on non-overlapping named entity recognition compared to sequence labeling models.

To tackle the aforementioned drawbacks, we propose a novel neural framework, named {\ModelName}, for Multi-Grained Named Entity Recognition. It is suitable for tackling both Nested NER and Non-overlapping NER.
The idea of {\ModelName} is natural and intuitive, which is to first detect entity positions in various granularities via a Detector and then classify these entities into different pre-defined categories via a Classifier. {\ModelName} has five types of modules: Word Processor, Sentence Processor, Entity Processor, Detection Network, and Classification Network, where each module can adopt a wide range of neural network designs.

In summary, the contributions of this work are:
\begin{itemize}[leftmargin=*]
    \setlength\itemsep{0mm}
    \item We propose a novel neural framework named {\ModelName} for Multi-Grained Named Entity Recognition, aiming to detect both nested and non-overlapping named entities effectively in a single model.
    
    \item {\ModelName} is highly modularized. 
    Each module in {\ModelName} can adopt a wide range of neural network designs. Moreover, {\ModelName} can be easily extended to many other related information extraction tasks, such as chunking \cite{ramshaw1999text} and slot filling \cite{mesnil2015using}.
    
    \item Experimental results show that {\ModelName} is able to achieve new state-of-the-art results on both Nested Named Entity Recognition tasks and Non-overlapping Named Entity Recognition tasks.
    
\end{itemize}

%% file: 2related.tex
\begin{figure*}[!ht]
    \centering
    \includegraphics[width=0.8\linewidth]{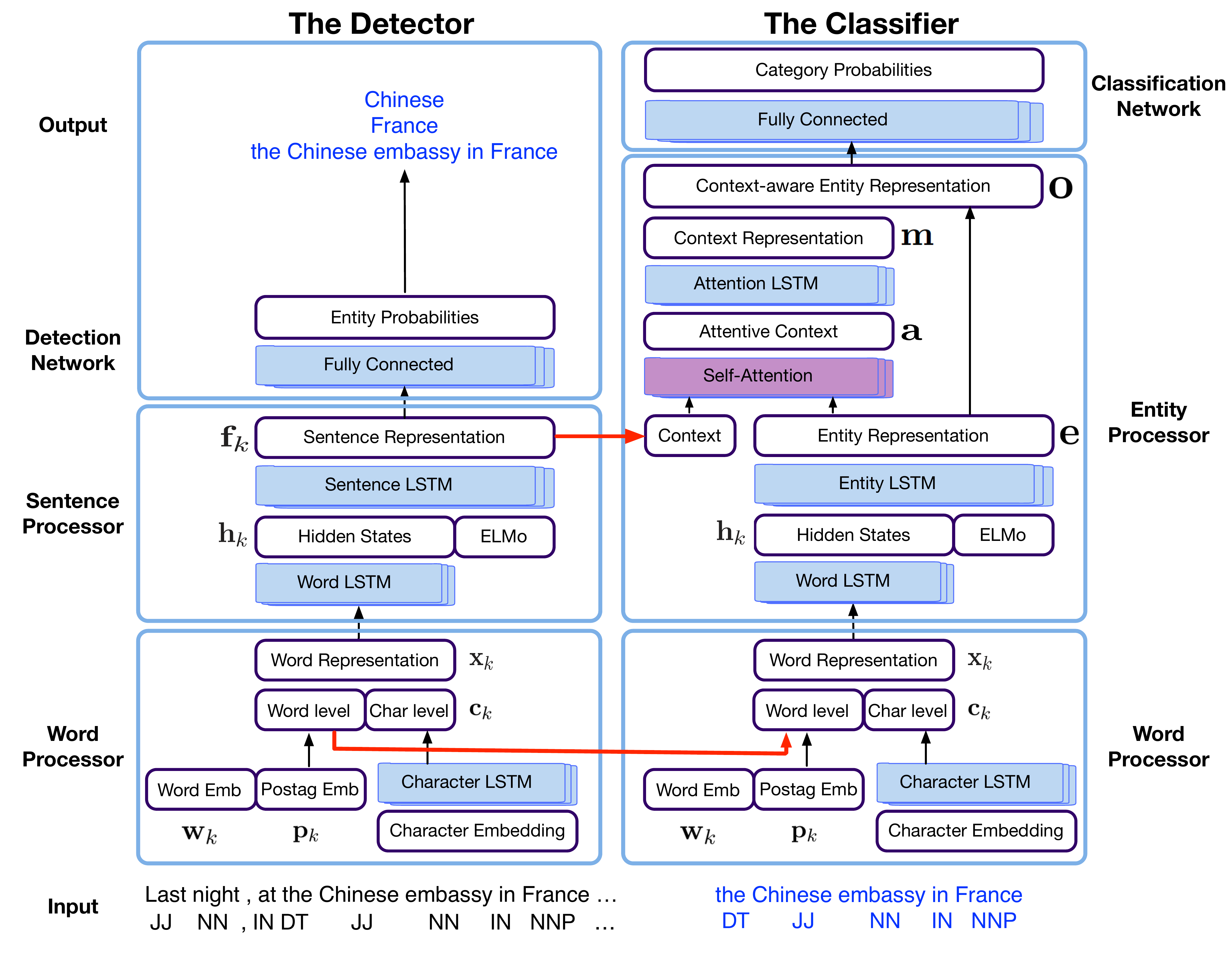}
    \caption{The framework of {\ModelName} for {\multi} Named Entity Recognition. It consists of a Detector and a Classifier.
    }
    \label{fig:EPN}
\end{figure*}

\section{Related Work}
Existing approaches for recognizing non-overlapping named entities usually treat the NER task as a sequence labeling problem. Various sequence labeling models achieve decent performance on NER, 
including probabilistic graph models such as Conditional Random Fields (CRF) \cite{ratinov2009design}, and deep neural networks like recurrent neural networks or convolutional neural networks (CNN). \citet{hammerton2003named} is the first work to use Long Short-Term Memory (LSTM) for NER.
\citet{collobert2011natural} employ a CNN-CRF structure, which obtains competitive results to statistical models. 
Most recent works leverage an LSTM-CRF architecture.
\citet{huang2015bidirectional} use hand-crafted spelling features; \citet{ma2016end} and \citet{chiu2016named} utilize a character CNN to represent spelling characteristics; \citet{lample2016neural} employ a character LSTM instead. Moreover, the attention mechanism is also introduced in NER to dynamically decide how much information to use from a word or character level component \cite{rei2016attending}.

External resources have been used to further improve the NER performance.
\citet{peters2017semi} add pre-trained context embeddings from bidirectional language models to NER.
\citet{peters2018deep} learn a linear combination of internal hidden states stacked in a deep bidirectional language model, ELMo, to utilize both higher-level states which capture context-dependent aspects and lower-level states which model aspects of syntax. 
These sequence labeling models can only detect non-overlapping entities and fail to detect nested ones.

Various approaches have been proposed for Nested Named Entity Recognition.
\citet{finkel2009nested} propose a CRF-based constituency parser which takes each named entity as a constituent in the parsing tree.
\citet{ju2018neural} dynamically stack multiple flat NER layers and extract outer entities based on the inner ones. Such model may suffer from the error propagation problem if shorter entities are recognized incorrectly.

Another series of approaches for Nested NER are based on hypergraphs.
The idea of using hypergraph is first introduced in \citet{lu2015joint}, which allows edges to be connected to different types of nodes to represent nested entities.
\citet{lu2017mention} use a multigraph representation and introduce the notion of mention separator for nested entity detection.
Both \citet{lu2015joint} and \citet{lu2017mention} rely on the hand-crafted features to extract nested entities and suffer from structural ambiguity issue. \citet{wang2018neural} present a neural segmental hypergraph model using neural networks to obtain distributed feature representation.
\citet{katiyar2018nested} also adopt a hypergraph-based formulation and learn the structure using an LSTM network in a greedy manner. 
One issue of these hypergraph approaches is the spurious structures of hypergraphs as they enumerate combinations of nodes, types and boundaries to represent entities. In other words, these models are specially designed for the nested named entities and are not suitable for the non-overlapping named entity recognition.

\citet{xu2017local} propose a local detection method which relies on a Fixed-size Ordinally Forgetting Encoding (FOFE) method to encode utterance and a simple feed-forward neural network to either reject or predict the entity label for each individual text fragment \cite{luan2018multi, lee2017end, he2018jointly}. Their model is in the same track with the framework we proposed whereas the difference is that we separate the NER task into two stages, {\it i.e.}, detecting entity positions and classifying entity categories.

%% file: 3model.tex
\section{The Proposed Framework}
An overview of the proposed {\ModelName} framework for multi-grained entity recognition, is illustrated in Figure \ref{fig:EPN}.
Specifically, {\ModelName} consists of two sub-networks: the Detector and the Classifier. 
The Detector detects all the possible entity positions while the Classifier aims at classifying detected entities into pre-defined entity categories. 
The Detector has three modules: 1) Word Processor which extracts word-level semantic features, 2) Sentence Processor that learns context information for each utterance and 3) Detection Network that decides whether a word segment is an entity or not.
The Classifier consists of 1) Word Processor which has the same structure as the one in the Detector, 2) Entity Processor that obtains entity features and 3) Classification Network that classifies entity into pre-defined categories. 
In addition, a self-attention mechanism is adopted in the Entity Processor to help the model capture and utilize entity-related contextual information.

Each module in {\ModelName} can be replaced with a wide range of different neural network designs. For example, BERT \cite{devlin2018bert} can be used as the Word Processor and a capsule model \cite{sabour2017dynamic, xia2018zero} can be integrated into the Classification Network. 

It is worth mentioning that in order to improve the learning speed as well as the performance of {\ModelName}, the Detector and the Classifier are trained with a series of shared input features, including the pre-trained word embeddings and the pre-trained language model features.
Sentence-level semantic features trained in the Detector are also transferred into the Classifier to introduce and utilize the contextual information.
We present the key building blocks and the properties of the Detector in Section 3.1 and the Classifier in Section 3.2, respectively.

\subsection{The Detector}
The Detector is aimed at detecting possible entity positions within each utterance. It takes an utterance as the input and outputs a set of entity candidates.
Essentially, we use a semi-supervised neural network inspired by \cite{peters2017semi} to model this process. The architecture of the Detector is illustrated in the left part of Figure \ref{fig:EPN}. Three major modules are contained in the Detector: Word Processor, Sentence Processor and Detection Network. More specifically, pre-trained word embeddings, POS tag information and character-level word information are used for generating semantically meaningful word representations. Word representations obtained from the Word Processor and the language model embeddings---ELMo \cite{peters2018deep}, are concatenated together to produce context-aware sentence representations. Each possible word segment is then examined in the Detection Network and to be decided whether accepted it as an entity or not. 

\subsubsection{Word Processor}
Word Processor extracts semantically meaningful word representation for each token.
Given an input utterance with $K$ tokens $(t_{1}, ..., t_{K})$, each token $t_k (1\leq k \leq K)$ is represented as
\begin{equation*}
    \mathbf{x}_{k}= [ \mathbf{w}_{k}; \mathbf{p}_{k}; \mathbf{c}_{k}],
\end{equation*}\vspace{-1.5em}

\noindent by using a concatenation of a pre-trained word embedding $\mathbf{w}_{k}$, POS tag embedding $\mathbf{p}_{k}$ if it exists, and a character-level word information $\mathbf{c}_k$.
The pre-trained word embedding $\mathbf{w}_k$ with a dimension $D_w$ is obtained from GloVe \cite{pennington2014glove}.
The character-level word information $\mathbf{c}_k$ is obtained with a bidirectional LSTM \cite{hochreiter1997long} layer to capture the morphological information. The hidden size of this character LSTM is set as $D_{cl}$.
As shown in the bottom of Figure \ref{fig:EPN}, character embeddings are fed into the character LSTM.
The final hidden states from the forward and backward character LSTM are concatenated as the character-level word information $\mathbf{c}_k$.
Those POS tagging embeddings and character embeddings are randomly initialized and learned within the learning process.

\subsubsection{Sentence Processor}
To learn the contextual information from each sentence, another bidirectional LSTM, named word LSTM, is applied to sequentially encode the utterance. 
For each token, the forward hidden states ${\mathord{\buildrel{\lower3pt\hbox{$\scriptscriptstyle\rightarrow$}}\over {\mathbf{h}}} }_{k}$ and the backward hidden states ${\mathord{\buildrel{\lower3pt\hbox{$\scriptscriptstyle\leftarrow$}}\over {\mathbf{h}}} }_{k}$ are concatenated into the hidden states $\mathbf{h}_{k}$.
The dimension of the hidden states of the word LSTM is set as $D_{wl}$.
\begin{align}
\begin{split}
     {\mathord{\buildrel{\lower3pt\hbox{$\scriptscriptstyle\rightarrow$}}\over {\mathbf{h}}} }_{k}  &= {\rm LSTM}_{fw} (\mathbf{x}_{k}, {\mathord{\buildrel{\lower3pt\hbox{$\scriptscriptstyle\leftarrow$}}\over {\mathbf{h}}} }_{k-1} ), \\
     ~~{\mathord{\buildrel{\lower3pt\hbox{$\scriptscriptstyle\leftarrow$}}\over {\mathbf{h}}} }_{k} &= {\rm LSTM}_{bw} (\mathbf{x}_{k},  {\mathord{\buildrel{\lower3pt\hbox{$\scriptscriptstyle\leftarrow$}}\over {\mathbf{h}}} }_{k+1}), \\
     ~~\mathbf{h}_{k} &= [ {\mathord{\buildrel{\lower3pt\hbox{$\scriptscriptstyle\rightarrow$}}\over {\mathbf{h}}} }_{k}; {\mathord{\buildrel{\lower3pt\hbox{$\scriptscriptstyle\leftarrow$}}\over {\mathbf{h}}} }_{k}]. 
\end{split}
\end{align}
Besides, we also utilize the language model embeddings pre-trained in an unsupervised way as the ELMo model in \citep{peters2018deep}. The pre-trained ELMo embeddings and the hidden states in the word LSTM $\mathbf{h}_{k}$ are concatenated.
Hence, the concatenated hidden states $\mathbf{h}_{k}$ for each token can be reformulated as:
\begin{equation}
    \mathbf{h}_{k} = [ {\mathord{\buildrel{\lower3pt\hbox{$\scriptscriptstyle\rightarrow$}}\over {\mathbf{h}}} }_{k}; {\mathord{\buildrel{\lower3pt\hbox{$\scriptscriptstyle\leftarrow$}}\over {\mathbf{h}}} }_{k}; \mathbf{ELMo}_{k}],
\end{equation}
where $\mathbf{ELMo}_{k}$ is the ELMo embeddings for token $t_k$. Speficially, a three-layer bi-LSTM neural network is trained as the language model. Since the lower-level LSTM hidden states have the ability to model syntax properties and higher-level LSTM hidden states can capture contextual information, ELMo computes the language model embeddings as a weighted combination of all the bidirectional LSTM hidden states:
\begin{equation}
    \mathbf{ELMo}_{k} = \gamma \sum\nolimits_{l = 0}^L {{u_j}} \mathbf{h}_{k,l}^{LM},
\end{equation}
where $\gamma$ is a task-specified scale parameter which indicates the importance of the entire ELMo vector to the NER task. $L$ is the number of layers used in the pre-trained language model, the vector $\mathbf{u} = [u_0, \cdots, u_L]$ represents softmax-normalized weights that combine different layers. $\mathbf{h}_{k,l}^{LM}$ is the language model hidden state of layer $l$ at the time step $k$.

A sentence bidirectional LSTM layer with a hidden dimension of $D_{sl}$ is employed on top of the concatenated hidden states $\mathbf{h}_{k}$. The forward and backward hidden states in this sentence LSTM are concatenated for each token as the final sentence representation $\mathbf{f}_{k}$ $\in \mathbb{R}^{2D_{sl}}$.

\subsubsection{Detection Network}
Using the semantically meaningful features obtained in $\mathbf{f}_{k}$, we can identify possible entities within each utterance. The strategy of finding entities is to first generate all the word segments as entity proposals and then estimate the probability of each proposal as being an entity or not. 

\begin{figure}[htbp!]
    \centering
    \includegraphics[width=0.9\linewidth]{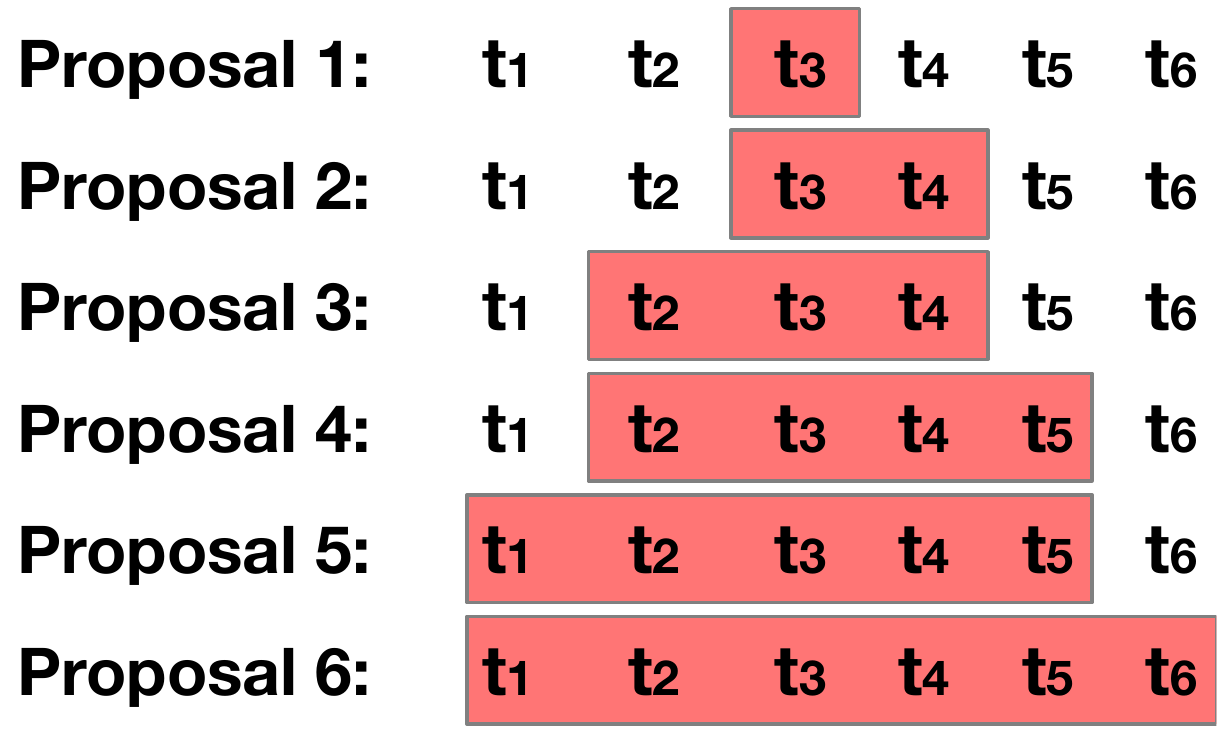}\vspace{-2mm}
    \caption{All possible entity proposals generated surrounding token $t_{3}$ when the maximum length of an entity proposal $R$ is set as 6.}
    \label{fig:entity_proposal}
\end{figure}

To enumerate all possible entity proposals, different lengths of entity proposals are generated surrounding each token position.
For each token position, R entity proposals with the length varies from 1 to the maximum length R are generated.
Specifically, it is assumed that an input utterance consists of a sequence of $N$ tokens $(t_{1}, t_{2}, t_{3}, t_{4}, t_{5}, t_{6}, ..., t_{N})$. 
To balance the performance and the computational cost, we set $R$ as 6.
We take each token position as the center and generate 6 proposals surrounding it. All the possible $6N$ proposals under the max-length of 6 will be generated.
As shown in Figure \ref{fig:entity_proposal}, the entity proposals generated surrounding token $t_3$ are: $(t_{3})$, $(t_{3}, t_{4})$, $(t_{2}, t_{3}, t_{4})$, $(t_{2}, t_{3}, t_{4}, t_{5})$, $(t_{1}, t_{2},  t_{3}, t_{4}, t_{5})$, $(t_{1}, t_{2},  t_{3}, t_{4}, t_{5}, t_{6})$. 
Similar entity proposals are generated for all the token positions and proposals that contain invalid indexes like ($t_{0}$,$t_{1}$,$t_{2}$) will be deleted. Hence we can obtain all the valid entity proposals under the condition that the max length is R.

For each token, we simultaneously estimate the probability of a proposal being an entity or not for R proposals.
A fully connected layer with a two-class softmax function is used to determine the quality of entity proposals:
\begin{equation}
    \mathbf{s}_{k} = {\rm softmax}\left (\mathbf{f}_{k}\mathbf{W}_{p} + \mathbf{b}_{p}\right ),
\end{equation}
where \( \mathbf{W}_{p} \in \mathbb{R}^{2D_{sl} \times 2R}\) and \( \mathbf{b}_{p} \in \mathbb{R}^{2R}\) are weights and the bias for the entity proposal layer; $\mathbf{s}_{k}$ contains $2R$ scores including $R$ scores for being an entity and R scores for not being an entity at position $k$. The cross-entropy loss is employed in the Detector as follows:
\begin{equation}
   {L_p} =  - \sum\nolimits_{k = 1}^K {\sum\nolimits_{r = 1}^R {{\bf{y}}_{k}^{r}\log {\bf{s}}_k^r} },
\end{equation}
where ${\bf{y}}_{k}^{r}$ is the label for proposal type $r$ at position $k$ and ${\bf{s}}_{k}^{r}$ is the probability of being an entity for proposal type $r$ at position $k$.
It is worth mentioning that, most entity proposals are negative proposals. Thus, 
to balance the influence of positive proposals and negative proposals in the loss function, we keep all positive proposals and use down-sampling for negative proposals when calculating the loss $L_p$.  
For each batch, we fix the number of the total proposals, including all positive proposals and sampled negative proposals, used in the loss function as $N_b$.
In the inference procedure of the Detection Network, an entity proposal will be recognized as an entity candidate if its score of being an entity is higher than score of not being an entity. 

\subsection{The Classifier}
The Classifier module aims at classifying entity candidates obtained from the Detector into different pre-defined entity categories. For the nested NER task, all the proposed entities will be saved and fed into the Classifier. For the NER task which has non-overlapping entities, we utilize the non-maximum suppression (NMS) algorithm \citep{neubeck2006efficient} to deal with redundant, overlapping entity proposals and output real entity candidates.
The idea of NMS is simple but effective: picking the entity proposal with the maximum probability, deleting conflict entity proposals, and repeating the previous process until all the proposals are processed. Eventually, we can get those non-conflict entity candidates as the input of the Classifier.

To understand the contextual information of the proposed entity, we utilize both sentence-level context information and a self-attention mechanism to help the model focus on entity-related context tokens. 
The framework of the Classifier is shown in the right part of Figure \ref{fig:EPN}. Essentially, it consists of three modules: Word Processor, Entity Processor and Classification Network.

\begin{table*}[!ht]
\centering
\resizebox{\linewidth}{!}{% <-
\begin{tabular}{ll|ccc|ccc|ccc}
 &&\multicolumn{3}{c|}{\textbf{ACE-2004}}  & \multicolumn{3}{c|}{\textbf{ACE-2005}} & \multicolumn{3}{c}{\textbf{CoNLL-2003}}   \\
&& \textbf{TRAIN}     & \textbf{DEV}   & \textbf{TEST}      & \textbf{TRAIN}     & \textbf{DEV}   & \textbf{TEST}      & \textbf{TRAIN}     & \textbf{DEV}   & \textbf{TEST}      \\
 \hline
\multirow{2}{*}{sentences}&\#total          & 6,799    &829     & 879          & 7,336     &958     & 1,047       &14,987   &3,466&3,684   \\
&\#overlaps               & 2,683(39\%) &293(35\%) & 373(42\%)   & 2,683 (37\%) &340(35\%) & 330 (32\%)  & - &-& - \\\hline
\multirow{4}{*}{entities}&\#total           & 22,207    &2,511     & 3,031        & 24,687    &3,217    & 3,027       & 23,499      &5,942 &5,648      \\
&\#overlaps                & 10,170 (46\%) & 1,091(43\%) & 1,418 (47\%) &        9,937 (40\%)    &1,192(37\%)   &    1,184 (39\%)         & - & - &- \\
&length \textgreater 6 & 1,439 (6\%) &179(7\%)  & 199 (7\%)    & 1,343 (5\%)  &148(5\%) & 160 (6\%)   & 23(0.1\%) &8(0.1\%) & 0 (0\%)    \\
&max length            & 57        &35    & 43           & 49      &30      & 27          & 10 &10     & 6     \\  
\end{tabular}
}
\caption{Corpora Statistics for the ACE-2004, ACE-2005 and CoNLL-2003 datasets.}
\label{data_statistics}
\end{table*}

\subsubsection{Word Processor}
A same Word Processor as in the Detector is used here to get the word representation for the entity candidates obtained from the Detector.
The word-level embedding, which is the concatenation of pre-trained word embedding and POS tag embedding if it is exists, is transferred from the Word Processor in the Detector to improve the performance as
well as to speed up the learning process. The character-level LSTM and character embeddings are trained separately in the Detector and the Classifier.

\subsubsection{Entity Processor}
The word representation is fed into a bidirectional word LSTM with hidden size $D_{wl}$ and the hidden states are concatenated with the ELMo language model embeddings as the entity features. 
A bidirectional LSTM with hidden size $D_{el}$ is applied to the entity feature to capture sequence information among the entity words. The last hidden states of the forward and backward Entity LSTM are concatenated as the entity representation $\mathbf{e} \in \mathbb{R}^{2D_{el}}$. 

The same word in different contexts may have different semantic meanings. 
To this end, in our model, we take the contextual information into consideration when learning the semantic representations of entity candidates. 
We capture the contextual information from other words in the same utterance. Denote $\mathbf{c}$ as the context feature vector for these context words, and it can be extracted from the sentence representation $\mathbf{f}_{k}$ in the Detector.
Hence, the sentence features trained in the Detector is directly transferred to the Classifier.

An easy way to model context words is to concatenate all the word representations or average them. However, this naive approach may fail when there exists a lot of unrelated context words. To select high-relevant context words and learn an accurate contextual representation, we propose a self-attention mechanism to simulate and dynamically control the relatedness between the context and the entity.
The self-attention module takes the entity representation $\mathbf{e}$ and all the context features $\mathbf{C} =  [\mathbf{c_{1}}, \mathbf{c_{2}}, ...,  \mathbf{c_{N}}]$ as the inputs, and outputs a vector of attention weights $\mathbf{a}$:
\begin{equation}
    \mathbf{a} = softmax( \mathbf{C} \mathbf{W} \mathbf{e}^T),
\end{equation}
where $\mathbf{W} \in \mathbb{R}^{2D_{sl} \times 2D_{el}}$ is a weight matrix for the self-attention layer, and $\mathbf{a}$ is the self-attention weight on different context words. To help the model focus on entity-related context, the attentive vector $\mathbf{C}^{att}$ is calculated as the attention-weighted context:
\begin{equation}
\mathbf{C}^{att} = \mathbf{a} * \mathbf{C}.
\end{equation}
The lengths of the attentive context $\mathbf{C}^{att}$ varies in different contexts. However, the goal of the Classification Network is to classify entity candidates into different categories, and thus it requires a fixed embedding size. We achieve that by adding another LSTM layer. An Attention LSTM with the hidden dimension $D_{ml}$ is used and the concatenation of the last hidden states in the forward and backward LSTM layer as the context representation $\mathbf{m} \in \mathbb{R}^{2D_{ml}}$. Hence the shape of the context representation is aligned.
We concatenate the context representation and the entity representation together as a context-aware entity representation to classify entity candidates: $ \mathbf{o} = [\mathbf{m}; \mathbf{e}]$.

\subsubsection{Classification Network}
A two-layer fully connected neural network is used to classify candidates into pre-defined categories:
\begin{equation}
   \mathbf{p}= {\rm softmax}\left({ \mathbf{W}_{c2}\left( {\sigma \left (\mathbf{o}\mathbf{W}_{c1} + \mathbf{b}_{c1}\right )}\right)}  + \mathbf{b}_{c2} \right),
\end{equation}
where $\mathbf{W_{c1}} \in \mathbb{R}^{(2D_{ml}+2D_{el}) \times D_h}$, $\mathbf{b_{c1}} \in \mathbb{R}^{D_h}$ , $\mathbf{W_{c2}} \in \mathbb{R}^{D_{c1} \times (D_{t}+1)}$, $\mathbf{b_{c2}} \in \mathbb{R}^{D_t+1}$ are the weights for this fully connected neural network, and $D_t$ is the number of entity types. Actually, this classification function classifies entity candidates into $(D_{t}+1)$ types. Here we add one more type as for the scenario that a candidate may not be a real entity.
Finally, the hinge-ranking loss is adopted in the Classification Network:
\begin{equation}
L_{c} =  \sum\nolimits_{y_w \in Y_w} max \left \{  0,\Delta +\mathbf{p}_{y_w} - \mathbf{p}_{y_r}  \right \},
\end{equation}
where $\mathbf{p}_{w}$ is the probability for the wrong labels $y_w$, $\mathbf{p}_{r}$ is the probability for the right label $y_r$, and $\Delta$ is a margin.
The hinge-rank loss urges the probability for the right label higher than the probability for the wrong labels and improves the classification performance.

%% file: 4exp.tex
\section{Experiments}
To show the ability and effectiveness of our proposed framework, {\ModelName}, for {\multi} Named Entity Recognition, we conduct the experiments on both Nested NER task and traditional non-overlapping NER task.

\subsection{Datasets}
We mainly evaluate our framework on ACE-2004 and ACE-2005 \cite{doddington2004automatic} with the same splits used by previous works \cite{luo2015joint, wang2018neural} for the nested NER task. Specifically, seven different types of entities such as person, facility, weapon and vehicle, are contained in the ACE datasets.
For the traditional NER task, we use the CoNLL-2003 dataset \cite{tjong2003introduction} which contains four types of named entities: location, organization, person and miscellaneous. 
An overview of these three datasets is illustrated in Table \ref{data_statistics}. It can be observed that most entities are less or equal to 6 tokens, and thus we select the maximum entity length $R=6$.

\subsection{Implementation Details}
We performed random search \cite{bergstra2012random} for hyper-parameter optimization and selected the best setting based on performance on the development set.
We employ the Adam optimizer \cite{kingma2014adam} with learning rate decay for all the experiments. The learning rate is set as 0.001 at the beginning and exponential decayed by 0.9 after each epoch.
The batch size of utterances is set as 20. In order to balance the influence of positive proposals and negative proposals, we use down-sampling for negative ones and the total proposal number $N_b$ for each batch is 128.
To alleviate over-fitting, we add dropout regularizations after the word representation layer and all the LSTM layers with a dropout rate of 0.5.
In addition, we employ the early stopping strategy when there is no performance improvement on the development dataset after three epochs.
The pre-trained word embeddings are from GloVe \cite{pennington2014glove}, and the word embedding dimension $D_w$ is 300. Besides, the ELMo 5.5B data\footnote{https://allennlp.org/elmo} is utilized in the experiment for the language model embedding. 
Moreover, the size of character embedding  $\mathbf{c}_k$ is 100, and the hidden size of the Character LSTM $D_{cl}$ is also 100.
The size of POS tag embedding $\mathbf{p}_{k}$ is 300 for the ACE datassets and no POS tag information is used in the CoNLL-2003 dataset. 
The hidden dimensions of the Word LSTM layer $D_{wl}$, the Sentence LSTM layer $D_{sl}$, the Entity LSTM layer $D_{el}$ and the Attention LSTM layer $D_{ml}$ are all set to 300. 
The hidden dimension of the classification layer $D_h$ is 50. The margin $\Delta$ in the hinge-ranking loss for the entity category classification is set to 5.
The ELMo scale parameter $\gamma$ used in the Detector is 3.35 and 3.05 in the Classifier, respectively.

\begin{table}[htbp!]
\centering
\resizebox{\linewidth}{!}{% <-
\begin{tabular}{l|ccc|ccc}
\multirow{2}{*}{\textbf{MODEL}} & \multicolumn{3}{c|}{\textbf{ACE-2004}} & \multicolumn{3}{c}{\textbf{ACE-2005}}  \\ 
                  & \textbf{P}       & \textbf{R}         & \textbf{F1}   & \textbf{P}          & \textbf{R}         & \textbf{F1}            \\\hline
\citet{lu2015joint}&70.0&56.9&62.8&66.3&59.2 &62.5\\
\citet{lample2016neural}& 71.3 &50.5 & 58.3&   64.1 & 52.4& 57.6  \\
\citet{lu2017mention}&72.7&58.0&64.5
&69.1&58.1&63.1\\
\citet{xu2017local}    &68.2&54.3&60.5&67.4 & 55.1 & 60.6  \\
\citet{katiyar2018nested} & 73.6 &71.8&72.7 &70.6 & 70.4 & 70.5  \\
\citet{ju2018neural}  &-&-&-&74.2&70.3&72.2 \\
\citet{wang2018neuralshort}&74.9 & 71.8& 73.3
&74.5 &71.5 &73.0 \\  
\citet{wang2018neural}  & 78.0& 72.4 &75.1          & 76.8     & 72.3   & 74.5  \\ \hline
{\ModelName} w/o context &79.8&76.3&78.0& \textbf{79.6}&75.6&77.5\\
{\ModelName} w/o attention &81.5&76.5&78.9& 79.4 &76.0 & 77.7 \\
{\ModelName} &\textbf{81.7}&\textbf{77.4}&\textbf{79.5}&79.0&\textbf{77.3}&\textbf{78.2}\\
\end{tabular}
}
\caption{Performance on ACE-2004 and ACE-2005 test set for the Nested NER task.}
\label{multi_result}
\end{table}

\subsection{Results}
{\bf Nested NER Task}.
The proposed {\ModelName} is very suitable for detecting nested named entities since every possible entity will be examined and classified. In order to validate this advantage, we compare {\ModelName} with numerous baseline models: 1) \citet{lu2015joint} which propose the mention hypergraphs for recognizing overlapping entities; 2) \citet{lample2016neural} which adopt the LSTM-CRF stucture for sequence labelling; 3) \citet{lu2017mention} which introduce mention separators to tag gaps between words for recognizing overlapping mentions; 4) \citet{xu2017local} that propose a local detection method; 5) \citet{katiyar2018nested} which propose a hypergraph-based model using LSTM for learning feature representations; 6) \citet{ju2018neural} that use a layered model which extracts outer entities based on inner ones; 7) \citet{wang2018neuralshort} which propose a neural transition-based model that constructs nested mentions through a sequence of actions; 8) \citet{wang2018neural} which adopt a neural segmental hypergraph model.

Experiment results of the Nested NER task on the ACE-2004 and ACE-2005 datasets are reported in Table \ref{multi_result}. 
We can observe from Table \ref{multi_result} that, our proposed framework {\ModelName} outperforms all the baseline approaches.
For both datasets, our model improves the state-of-the-art result by around 4\% in terms of precision, recall, as well as the F1 score.

To study the contribution of different modules in {\ModelName}, we also report the performance of two ablation variations of the proposed {\ModelName} at the bottom of Table \ref{multi_result}.
{\ModelName} w/o attention is a variation of {\ModelName} which removes the self-attention mechanism and {\ModelName} w/o context removes all the context information.
To remove the self-attention mechanism, we feed the context feature $\mathbf{C}$ directly into a bi-directional LSTM to obtain context representation $\mathbf{m}$, other than the attentive context vector $\mathbf{C}^{att}$. As for  {\ModelName} w/o context, we only use entity representation $\mathbf{e}$ to do classification other than the context-aware entity representation $\mathbf{o}$.
By adding the context information, the F1 score improves 0.9\% on the ACE-2004 dataset and 0.7\% on the ACE-2005 dataset. The self-attention mechanism improves the F1 score by 0.6\% on the ACE-2004 dataset and 0.5\% on the ACE-2005 dataset.

\begin{table}[htbp!]
\centering
\resizebox{\linewidth}{!}{% <-
\begin{tabular}{l|ccc|ccc}
\multirow{2}{*}{\textbf{MODEL}} & \multicolumn{3}{c|}{\textbf{OVERLAPPING}} & \multicolumn{3}{c}{\textbf{NON-OVERLAPPING}} \\
 & \textbf{P}        & \textbf{R}        & \textbf{F1}        & \textbf{P}          & \textbf{R}         & \textbf{F1}       \\ \hline
 \citet{lu2015joint}&   68.1       &     52.6     &  59.4        &   64.1        &  65.1 & 64.6          \\
\citet{lu2017mention} &     70.4     &    55.0      &     61.8      &   67.2         &     63.4      &       65.2     \\
 \citet{wang2018neuralshort} &     77.4    &    70.5      &    73.8       &      76.1    &   69.6        &     72.7       \\
 \citet{wang2018neural} &    80.6     &    73.6      &    76.9       &       75.5     &     71.5      &   73.4     \\ \hline
{\ModelName} &      \textbf{82.6}    &     \textbf{76.0}    &    \textbf{79.2}       &   \textbf{77.8}     &    \textbf{79.5}      &  \textbf{78.6}   \\
\end{tabular}
}
\caption{Results on different types of sentences (ACE-2005).}
\label{overlap}
\end{table}

To analyze how well our model performs on overlapping and non-overlapping entities, we split the test data into two portions: sentences with and without overlapping entities (follow the splits used by \citet{wang2018neural}). Four state-of-the-art nested NER models are compared with our proposed framework {\ModelName} on the ACE-2005 dataset. As illustrated in Table \ref{overlap}, {\ModelName} consistently performs better than the baselines on both portions, especially for the non-overlapping part. This observation indicates that our model can better recognize non-overlapping entities than previous nested NER models.

The first step in {\ModelName} is to detect entity positions using the Detector, where the effectiveness of proposing correct entity candidates immediately affects the performance of the whole model. To this end, we provide the experiment results of detecting correct entities in the Detector module here. The precision, recall and F1 score are 85.23 , 91.84, 88.41 for the ACE-2004 dataset and 84.95, 89.35, 87.09 for the ACE-2005 dataset.

\begin{table}[htbp!]
\centering
\resizebox{\linewidth}{!}{% <-
\begin{tabular}{l|cc}
\multicolumn{1}{l|}{\multirow{2}{*}{\textbf{MODEL}}} & \multicolumn{2}{c}{\textbf{CoNLL-2003}}  \\
\multicolumn{1}{c|}{}      & \textbf{DEV}      &\textbf{TEST} \\ \hline
\citet{lu2015joint} & 89.2 & 83.8 \\
\citet{lu2017mention} & - & 84.3 \\
\citet{xu2017local} &  -  & 90.85 \\
\citet{wang2018neural} &-&90.2\\\hline
\citet{lample2016neural}    &  -  &  90.94\\
\citet{ma2016end}   &94.74&  91.21\\ 
\citet{chiu2016named} &$94.03\pm 0.23$ & $91.62 \pm 0.33$ \\ 
\citet{peters2017semi}  & - & $91.93 \pm 0.19$\\
\citet{peters2018deep} &- & $92.22 \pm 0.10$ \\ 
\hline
{\ModelName} w/o context &$ 95.21 \pm 0.12$& $92.23 \pm 0.06 $\\
{\ModelName} w/o attention&$ 95.23 \pm 0.06$& $92.26 \pm 0.09 $\\
{\ModelName} &$ \textbf{95.24} \pm \textbf{0.13} $&$ \textbf{92.28} \pm \textbf{0.12} $\\
\end{tabular}
}
\caption{F1 scores on CoNLL-2003 devlopement set (DEV) and test set (TEST) for the English NER task. Mean and standard deviation across five runs are reported. Pos tags information are not used.}
\label{result}
\end{table}

{\bf NER Task}. We also evaluate the proposed {\ModelName} framework on the NER task which needs to reorganize non-overlapping entities. Two types of baseline models are compared here: sequence labelling models which are designed specifically for non-overlapping NER task and nested NER models which also provide the ability to detect non-overlapping mentions. The first type of models including 1) \citet{lample2016neural} which adopt the LSTM-CRF structure; 2) \citet{ma2016end} which use a LSTM-CNNs-CRF architecture; 3) \citet{chiu2016named} which propose a CNN-LSTM-CRF model; 4) \citet{peters2017semi} which add semi-supervised language model embeddings; and 5) \citet{peters2018deep} which utilize the state-of-the-art ELMo language model embeddings. The second types include four Nested models mentioned in the Nested NER section: 1) \citet{luo2015joint}; 2) \citet{lu2017mention}; 3) \citet{xu2017local}; 4) \cite{wang2018neural}.

Table~\ref{result} shows the F1 scores of different approaches on CoNLL-2003 devlopement set and test set for the English NER task. Mean and standard deviation across five runs are reported.
It can be observed from Table~\ref{result} that the proposed {\ModelName} model outperforms all the baselines. The models designed for non-overlapping entity detection usually performs better than Nested NER models for the NER task. Our proposed framework outperforms state-of-the-art results both on the NER and Nested NER task. \citet{xu2017local} is the best baseline model among the Nested models since it shares a similar idea of our proposed framework by individually examining each entity proposal.
From the ablation study, we can observe that by purely adding the context information, the F1 score on the CoNLL-2003 test set improves from 92.23 to 92.26, and by adding the attention mechanism, the F1 score improves to 92.28. 

We also provide the performance of detecting non-overlapping entities in the Detector here. The precision, recall and F1 score are 95.33, 95.69 and 95.51 on the CoNLL-2003 dataset.

%% file: 5con.tex
\vspace{-2mm}
\section{Conclusions}
\vspace{-2mm}
In this work, we propose a novel neural framework named {\ModelName} for {\multi} Named Entity Recognition where multiple entities or entity mentions in a sentence could be non-overlapping or totally nested. 
{\ModelName} is framework with high modularity and each component in {\ModelName} can adopt a wide range of neural networks. 
Experimental results show that {\ModelName} is able to achieve state-of-the-art results on both nested NER task and traditional non-overlapping NER task.

%% file: 6ack.tex
\section*{Acknowledgments}
We thank the reviewers for their valuable comments.
Special thanks go to Lu Wei from Singapore University of Technology and Design for sharing the datasets split details.
This work is supported in part by NSF through grants IIS-1526499, IIS-1763325, and CNS-1626432.